# Testing Identifiability of Causal Effects

**David Galles***
Cognitive Systems Laboratory
Computer Science Department
UCLA, Los Angeles, CA 90024
*galles@cs.ucla.edu*

**Judea Pearl**
Cognitive Systems Laboratory
Computer Science Department
UCLA, Los Angeles, CA 90024
*judea@cs.ucla.edu*

## Abstract

This paper concerns the probabilistic evaluation of the effects of actions in the presence of unmeasured variables. We show that the identification of causal effect between a singleton variable $X$ and a set of variables $Y$ can be accomplished systematically, in time polynomial in the number of variables in the graph. When the causal effect is identifiable, a closed-form expression can be obtained for the probability that the action will achieve a specified goal, or a set of goals.

## 1 Introduction

Assume we need to replace an expert operating a complex production plant. Before we take charge, we are given a blue print of the plant, together with explanations on the functions of the various dials and knobs, and we are able to observe the expert in actual operation over a long period of time. During this period, we record the dials which the expert consults prior to taking actions, we understand the function of those dials but we cannot observe the actual reading on each of those dials. The data we are able to collect during the observation period include: the actions taken by the agent, the readings of some of the dials, and the outcome of various performance indicators. Our problem is to predict, on the basis of the data thus collected, the effect of a given action on the performance of the plant.

The problem of learning from the performance of other agents is that one is never sure whether an observed response is due to the agent's action or due to events that triggered that action and simultaneously caused the response. Such events are called "confounders", and they present a major problem in the analysis of observational studies in the social and health sciences. For example, we cannot be sure if it was the drug that caused the patient to vomit, or the some prior condition which the doctor tried to treat by prescribing the drug. Similarly, we cannot tell whether higher taxes are responsible for the current recession or the economic indicators which government experts consulted before raising taxes.

The standard techniques of dealing with confounders is to adjust for possible variations in those environmental factors which might trigger the actions. This mounts to conditioning the observed distribution on the various levels of those factors, evaluating the action in each level separately, then taking the (weighted) average over those levels. However, in problems like the one describe above some of the confounding factors are unobservable, hence they cannot be conditioned on.

The techniques developed in this paper will enable us to recognize, by graphical means, whether a given action can be evaluated from joint distributions on observed quantities and, if the answer is positive, decide which quantities should be measured, and how to adjust for them. Technically speaking, the task accomplished parallels the identification of recursive structural equations in the presence of unmeasured variables. However, whereas traditional theories of identifiability deal exclusively with estimating coefficients in parametric equations, the identifiability problem solved in this paper is nonparametric; no assumptions are made regarding the functional forms of the structural equations or the distributions of the errors.[1]

Cast in terms of probabilistic networks, the problem addressed in this paper is the evaluation of the effects of an atomic action, $do(X = x)$, when the causal diagram is not fully specified. We are given the topology of the diagram but not the conditional probabilities on all variables. Numerical probabilities are given to only a subset of variables which are deemed "observable", while those deemed "unobservable" serve only to specify possible connections among observed quantities, but are not given numerical probabilities.

---

*David Galles is a graduate student

[1] Naturally, nonparametric identifiability is not concerned with values of numerical parameters but with the ultimate purpose to which parameters are being used in structural models, namely, the analysis of actions and causal effects.

The semantics behind causal diagrams and their relations to actions and belief networks have been discussed in prior publications [Pearl and Verma, 1991, Goldszmidt and Pearl, 1992, Druzdel and Simon, 1993, Pearl 1993, Spirtes *et al.*, 1993, Pearl 1994a]. In [Spirtes *et al*] for example, it was shown how causal networks can be used to facilitate quantitative predictions of the effects of interventions, including interventions that were not contemplated during the network's construction.[2] A previous UAI paper [Pearl, 1994b] reviews this aspect of causal networks, and proposes a calculus for deriving probabilistic assessments of the effects of actions in the presence of unmeasured variables. This calculus can be used to check or search for a proof that the effect of one variable on another is *identifiable*, namely, that it is possible to obtain a consistent estimate of the probability of $Y$ under the condition that $X$ is set to $x$ by external intervention, from data involving only observed variables. This paper systematizes the search for such a proof.

We provide a polynomial time graph-based method of determining whether the effect of one variable on another is identifiable.[3] If identifiability is confirmed, the method can provide closed-form expressions for the distribution of the outcome variable $Y$ under the external manipulation of the control variable $X$. The derived expression, denoted, $P(y|do(x))$, invokes only measured probabilities as obtained, for example, by recording past performance of other acting agents. Although the actions of those agents may have been triggered by hidden factors, unseen by the analyst, the impact of $X$ on $Y$ can still be predicted consistently, using observed variables only. If $Y$ stands for a goal variable, then the probability of reaching the goal through each action $do(X = x)$ can be determined consistently from such partial observations.

## 2    Notation and Technical Background

### 2.1    Causal theories, actions, causal effect, and identifiability

**Definition 1** *A causal theory is a 4-tuple*
$$T =< V, U, P(\mathbf{u}), \{f_i\} >$$
*where*

(i) $V = \{X_1, \ldots, X_n\}$ *is a set of observed variables*

(ii) $U = \{U_1, \ldots, U_m\}$ *is a set of unobserved variables which represent disturbances, abnormalities or assumptions,*

(iii) $P(\mathbf{u})$ *is a distribution function over $U_1, \ldots, U_m$, and*

---

[2]In contrast to influence diagrams which, lacking causal interpretation, require that action be contemplated in advance and incorporated as decision nodes in the diagrams.

[3]An extension of our analysis to the case of multiple actions (sequential or concurrent) is reported in [Pearl and Robins, 1995]

(iv) $\{f_i\}$ *is a set of n deterministic functions, each of the form*
$$X_i = f_i(X_1, \ldots, X_n, U_1, \ldots, U_m) \quad i = 1, \ldots, n \tag{1}$$

*We will assume that the set of equations in (iv) has a unique solution for $X_i, \ldots, X_n$, given any value of the disturbances $U_1, \ldots, U_m$. Therefore the distribution $P(\mathbf{u})$ induces a unique distribution on the observables, which we denote by $P_T(\mathbf{v})$.*

We will consider concurrent actions of the form $do(X = x)$, where $X \subseteq V$ is a set of variables and $x$ is a set of values from the domain of $X$. In other words, $do(X = x)$ represents a combination of actions that forces the variables in $X$ to attain the values $x$.

**Definition 2** *(Effect of actions) The effect of the action $do(X = x)$ on a causal theory $T$ is given by a subtheory $T_x$ of $T$, where $T_x$ obtains by deleting from $T$ all equations corresponding to variables in $X$ and substituting the equations $X = x$ instead.*

**Definition 3** *(causal effect) Given two disjoint subsets of variables, $X \subseteq V$ and $Y \subseteq V$, the causal effect of $X$ on $Y$, denoted $P_T(y|\hat{x})$, is a function from the domain of $X$ to the space of probability distributions on $Y$, such that*
$$P_T(y|\hat{x}) = P_{T_x}(y) \tag{2}$$
*for each realization $x$ of $X$. In other words for each $x \in dom(X)$, the causal effect $P_T(y|\hat{x})$ gives the distribution of $Y$ induced by the action $do(X = x)$.*

Note that causal effects are defined relative to a given causal theory $T$, though the subscript $T$ is often suppressed for brevity.

**Definition 4** *(identifiability) The causal effect of $X$ on $Y$ is said to be* identifiable *if the quantity $P(y|\hat{x})$ can be computed uniquely from any positive distribution of the observed variables, that is, if for every pair of theories $T_1$ an $T_2$ such that $P_{T_1}(\mathbf{v}) = P_{T_2}(\mathbf{v}) > 0$, we have $P_{T_1}(y|\hat{x}) = P_{T_2}(y|\hat{x})$*

Identifiability means that $P(y|\hat{x})$ can be estimated consistently from an arbitrarily large sample randomly drawn from the distribution of the observed variables.

**Definition 5** *(back-door path) A path from $X$ to $Y$ in a graph $G$ is said to be a* back-door path *if it contains an arrow into $X$.*

Figure 1 illustrates a simple causal theory in the form of a diagram. It describes the causal relationships among the season of the year $(X_1)$, whether rain falls $(X_2)$, whether the sprinkler is on $(X_3)$, whether the pavement would get wet $(X_4)$, and whether the pavement would be slippery $(X_5)$. All variables in this figure are binary, taking a value of either true or false, except the root variable $X_1$ which can take one of four values: Spring, Summer, Fall, or Winter. Here, the absence of a direct link between $X_1$ and $X_5$, for example, captures our understanding that the influence of seasonal variations on the slipperiness of the pavement is mediated by other conditions (e.g., the wetness of the pavement).

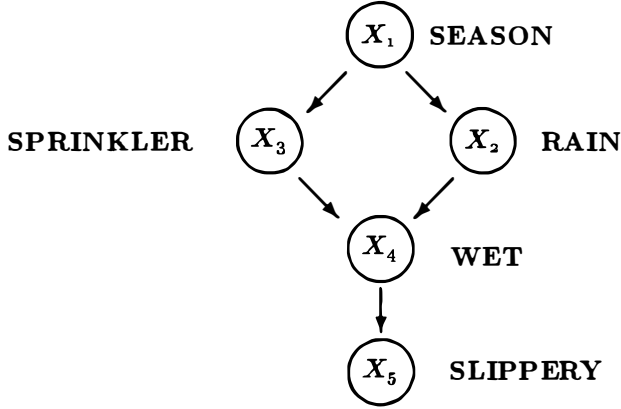

Figure 1: A diagram representing a causal theory on five variables.

The theory corresponding to Figure 1 consists of five functions, each representing an autonomous mechanism:

$$\begin{array}{lll} X_1 = U_1 & X_4 = f_4(X_3, X_2, U_4) \\ X_2 = f_2(X_1, U_2) & X_5 = f_5(X_4, U_5) \\ X_3 = f_3(X_1, U_3) & \end{array} \quad (3)$$

To represent the action "turning the sprinkler ON", $do(X_3 = \text{ON})$, we delete the equation $X_3 = f_3(x_1, u_3)$ from the theory of Eq. (3), and replace it with $X_3 = \text{ON}$. The resulting subtheory, $T_{X_3=\text{ON}}$, contains all the information needed for computing the effect of the actions on other variables. For example, it is easy to see from this subtheory that the only variables affected by the action are $X_4$ and $X_5$, that is, the descendant of the manipulated variable $X_3$.

The probabilistic analysis of causal theories becomes particularly simple when two conditions are satisfied:

1. The theory is recursive, i.e., there exists an ordering of the variables $V = \{X_1, \ldots, X_n\}$ such that each $X_i$ is a function of a subset $\mathbf{pa}_i$ of its predecessors

$$X_i = f_i(\mathbf{pa}_i, U_i), \quad \mathbf{pa}_i \subseteq \{X_1, \ldots, X_{i-1}\} \quad (4)$$

2. The disturbances $U_1, \ldots, U_n$ are mutually independent, $U_i \perp\!\!\!\perp U_j$, which also implies (from the exogeneity of the $U_i$'s)

$$U_i \perp\!\!\!\perp \{X_1, \ldots, X_{i-1}\} \quad (5)$$

These two conditions, also called Markovian, are the basis of Bayesian networks [Pearl, 1988] and they enable us to compute causal effects directly from the conditional probabilities $P(x_i|\mathbf{pa}_i)$, without specifying the functional form of the functions $f_i$, or the distributions $P(u_i)$ of the disturbances. This is seen immediately from the following observations:

The distribution induced by any Markovian theory $T$ is given by the product

$$P_T(x_1, \ldots, x_n) = \prod_i P(x_i|\mathbf{pa}_i) \quad (6)$$

where $\mathbf{pa}_i$ are the direct predecessors (called *parents*) of $X_i$ in the diagram. On the other hand the distribution induced by the The subtheory $T_{x'_j}$, representing the action $do(X_j = x'_j)$ is also Markovian, hence it also induces a product-like distribution

$$P_{T_{x'_j}}(x_1, \ldots, x_n)$$
$$= \begin{cases} \prod_{i \neq j} P(x_i|\mathbf{pa}_i) = \frac{P(x_1,\ldots,x_n)}{P(x_j|\mathbf{pa}_j)} & \text{if } x_j = x'_j \\ 0 & \text{if } x_j \neq x'_j \end{cases} \quad (7)$$

where the partial product reflects the surgical removal of

$$X_j = f_j(\mathbf{pa}_j, U_j)$$

from the theory of Eq. (4).

In the example of Figure 1, the pre-action distribution is given by the product

$$P_T(x_1, x_2, x_3, x_4, x_5)$$
$$= P(x_1)P(x_2|x_1)P(x_3|x_1)P(x_4|x_2, x_3)P(x_5|x_4) \quad (8)$$

while the surgery corresponding to the action $do(X_3 = \text{ON})$ amounts to deleting the link $X_1 \to X_3$ from the graph and fixing the value of $X_3$ to ON, yielding the post-action distribution:

$$P_T(x_1, x_2, x_4, x_5|do(X_3 = \text{ON}))$$
$$= P(x_1)\ P(x_2|x_1)\ P(x_4|x_2, X_3 = \text{ON})\ P(x_5|x_4) \quad (9)$$

Note the difference between the action $do(X_3 = \text{ON})$ and the observation $X_3 = \text{ON}$. The latter is encoded by ordinary Bayesian conditioning, while the former by conditioning a mutilated graph, with the link $X_1 \to X_3$ removed. This mirrors indeed the difference between seeing and doing: after observing that the sprinkler is ON, we wish to infer that the season is dry, that it probably did not rain, and so on; no such inferences should be drawn in evaluating the effects of the deliberate action "turning the sprinkler ON". The amputation of $X_3 = f_3(X_1, U_3)$ from (3) ensures the suppression of any abductive inferences from any of the action's consequences.

## 2.2 Action Calculus

The identifiability of causal effects demonstrated in Section 2.1 relies critically on the Markovian assumptions (4) and (5). If a variable that has two descendants in the graph is unobserved, the disturbances in the two equations are no longer independent, the Markovian property (4) is violated and identifiability may be destroyed. This can be seen easily from Eq. (7); if any parent of the manipulated variable $X_j$ is unobserved, one cannot estimate the conditional probability $P(x_j|\mathbf{pa}_j)$, and the effect of the action $do(X_j = x_j)$ may not be predictable from the observed distribution $P(x_1, \ldots, x_n)$. Fortunately, certain causal effects are identifiable even in situations where members of $\mathbf{pa}_j$ are be unobservable, and these situations can be recognized through the action calculus introduced in [Pearl, 1994a].

Let $X, Y$, and $Z$ be arbitrary disjoint sets of nodes in a DAG $G$. We say that $X$ and $Y$ are independent given $Z$ in $G$, denoted $(X \parallel Y|Z)_G$, if the set $Z$ $d$-separates $X$ from $Y$ in $G$. We denote by $G_{\overline{X}}$ the graph obtained by deleting from $G$ all arrows pointing to nodes in $X$. Likewise, we denote by $G_{\underline{X}}$ the graph obtained by deleting from $G$ all arrows emerging from nodes in $X$. To represent the deletion of both incoming and outgoing arrows, we use the notation $G_{\overline{X}\underline{Z}}$. Finally, the expression $P(y|\hat{x}, z) \triangleq P(y, z|\hat{x})/P(z|\hat{x})$ stands for the probability of $Y = y$ given that $Z = z$ is observed and $X$ is held constant at $x$.

**Theorem 1** Let $G$ be the directed acyclic graph associated with a Markovian causal theory, and let $P(\cdot)$ stand for the probability distribution induced by that theory. For any disjoint subsets of variables $X, Y, Z$, and $W$ we have:

**Rule 1** Insertion/deletion of observations

$$P(y|\hat{x}, z, w) = P(y|\hat{x}, w) \text{ if } (Y \parallel Z|X, W)_{G_{\overline{X}}} \quad (10)$$

**Rule 2** Action/observation exchange

$$P(y|\hat{x}, \hat{z}, w) = P(y|\hat{x}, z, w) \text{ if } (Y \parallel Z|X, W)_{G_{\overline{X}\underline{Z}}} \quad (11)$$

**Rule 3** Insertion/deletion of actions

$$P(y|\hat{x}, \hat{z}, w) = P(y|\hat{x}, w) \text{ if } (Y \parallel Z|X, W)_{G_{\overline{X}, \overline{Z(W)}}} \quad (12)$$

where $Z(W)$ is the set of $Z$-nodes that are not ancestors of any $W$-node in $G_{\overline{X}}$.

Each of the inference rules above follows from the basic interpretation of the "$\hat{x}$" operator as a replacement of the causal mechanism that connects $X$ to its pre-action parents by a new mechanism $X = x$ introduced by the intervening force. The result is a submodel characterized by the subgraph $G_{\overline{X}}$ (named "manipulated graph" in [Spirtes et al. 1993]) which supports all three rules.

**Corollary 1** A causal effect $q: P(y_1, \ldots, y_k|\hat{x}_1, \ldots, \hat{x}_m)$ is identifiable in a model characterized by a graph $G$ if there exists a finite sequence of transformations, each conforming to one of the inference rules in Theorem 1, which reduces $q$ into a standard (i.e., hat-free) probability expression involving observed quantities.    □

Although Theorem 1 and Corollary 1 require the Markovian property, they can also be applied to recursive non Markovian theories, because such theories become Markovian if we consider the unobserved variables as part of the analysis, and represent them as nodes in the graph. To demonstrate, assume that variable $X_1$ in Figure 1 is unobserved, rendering the disturbances $U_3$ and $U_2$ dependent, since these terms now include the common influence of $X_1$. Theorem 1 tells us that the causal effect $P(x_4|\hat{x}_3)$ is identifiable, because:

$$P(x_4|\hat{x}_3) = \sum_{x_1} P(x_4|\hat{x}_3, x_2)P(x_2|\hat{x}_3)$$

Rule 3 permits the reduction

$$P(x_2|\hat{x}_3) = P(x_2)$$

while Rule 2 permits the exchange

$$P(x_4|\hat{x}_3, x_2) = P(x_4|x_3, x_2)$$

This gives

$$P(x_4|\hat{x}_3) = \sum_{x_2} P(x_4|x_3, x_2)P(x_2)$$

which is a "hat-free" expression, involving only observed quantities.

In the sequel we will represent confounding unmeasured variables by dashed arcs. For example, the arc between $X$ and $B_2$ in Figure 2 stands for an unmeasured variable variable $U$ that is a common cause of $X$ and $B_2$.

## 3  A Graphical Criterion for Testing Identifiability

To avoid excessive notation, the rest of the paper will consistently refer to queries $P(y|\hat{x})$ that satisfies Corrolary 1 as "identifiable" with the understanding that it represents sufficient but not (yet) neccessary condition for semantical identifiability as in Definition 4. The two notions would be equivalent if the rules in Theorem 1 where complete.

**Theorem 2** *A necessary and sufficient condition for the identifiability of $P(y|\hat{x})$ in a graph $G$ is that $G$ satisfies one of the following four conditions :*

1. There is no directed path from $X$ to $Y$ in $G$.

2. There is no back-door path from $X$ to $Y$ in $G$, that is $(X \parallel Y)_{G_{\underline{X}}}$.

3. There exists a set of nodes $B$ that blocks all back-door paths from $X$ to $Y$ such that $P(b|\hat{x})$ is identifiable. (A special case of this condition occurs when $B$ consists entirely of non-descendants of $X$, in which case $P(b|\hat{x})$ reduces immediately to $P(b)$).

4. There exist sets of nodes $Z_1$ and $Z_2$ such that :
   - No element of $Z_2$ is a descendant of $X$
   - $Z_1$ blocks every directed path from $X$ to $Y$, ie., $(Y \perp\!\!\!\perp X|Z_1)_{G_{\overline{Z_1}\,\overline{X}}}$
   - $Z_2$ blocks all back-door paths between $Z_1$ and $Y$ in $G_{\overline{X}}$, ie., $(Y \perp\!\!\!\perp Z_1|Z_2)_{G_{\overline{X}\underline{Z_1}}}$,
   - $Z_2$ blocks all back-door paths between $X$ and $Z_1$, ie., $(X \perp\!\!\!\perp Z_1|Z_2)_{G_{\underline{X}}}$,

A special case of Condition 4 occurs when $Z_2 = \emptyset$ and there is no back door path from $X$ to $Z_1$ or from $Z_1$ to $Y$.

**Proof of Theorem 2**

We first prove the sufficiency of the four conditions above, then turn to prove their necessity.

- **Condition 1**: If there is no directed path from $X$ to $Y$ in $G$, then $(Y \perp\!\!\!\perp X)_{G_{\overline{X}}}$. So, by Rule 3, $P(y|\hat{x}) = P(y)$, and the query is identifiable.

- **Condition 2**: This follows directly from Rule 1 : If $(Y \perp\!\!\!\perp X)_{G_{\underline{X}}}$, then we can immediately change $P(y|\hat{x})$ to $P(y|x)$. Thus, the query is identifiable.

- **Condition 3**: If there is a set of nodes $B$ that blocks all back-door paths from $X$ to $Y$, then we can rewrite $P(y|\hat{x})$ as $\sum_b P(y|\hat{x}, b)P(b|\hat{x})$. Since $B$ blocks all back-door paths from $X$ to $Y$, it must be the case that $(Y \perp\!\!\!\perp X|B)_{G_{\underline{X}}}$, so, by Rule 2, we can rewrite $P(y|\hat{x}, b)$ as $P(y|x, b)$. If the query $(b|\hat{x})$ is identifiable, then the original query must also be identifiable. See Figure 2 for examples.

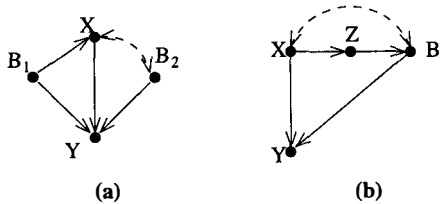

(a)          (b)

Figure 2: Illustrating Condition 3 of Theorem 2. In **a**, the set $\{B_1, B_2\}$ blocks all back-door paths from $X$ to $Y$ and $P(b_1, b_2|\hat{x}) = P(b_1, b_2)$. In **b**, the node $B$ blocks all back-door paths from $X$ to $Y$, and $P(b|\hat{x})$ is identifiable using Condition 4.

- **Condition 4**: (See Figure 3 for examples). If there is a set of nodes $Z_1$ that block all directed paths from $X$ to $Y$, and a set of nodes $Z_2$ that block all back-door paths between $Y$ and $Z_1$ in $G_{\overline{X}}$, expanding $P(y|\hat{x}) =$ $\sum_{z_1, z_2} P(y|\hat{x}, z_1, z_2) P(z_1, z_2|\hat{x})$. we can rewrite $P(y|\hat{x}, z_1, z_2)$ as $P(y|\hat{x}, \hat{z}_1, z_2)$ using Rule 2, since all back-door paths between $Z_1$ and $Y$ are blocked by $Z_2$ in $G_{\overline{X}}$. We can reduce $P(y|\hat{x}, \hat{z}_1, z_2)$ to $P(y|\hat{z}_1, z_2)$ using Rule 3, since $(Y \perp\!\!\!\perp X|Z_1, Z_2)_{G_{\overline{Z_1}\,\overline{X(Z_2)}}}$. We can rewrite $P(y|\hat{z}_1, z_2)$ as $P(y|z_1, z_2)$ if $(Y \perp\!\!\!\perp Z_1|Z_2)_{G_{\underline{Z_1}}}$. The only way that this independence cannot hold is if there is a path from $Y$ to $Z_1$ through $X$, since $(Y \perp\!\!\!\perp Z_1|Z_2)_{G_{\overline{X}\underline{Z_1}}}$. However, we can block this path by conditioning and summing over $X$ to get $\sum_{x'} P(y|\hat{z}_1, z_2, x') P(x'|\hat{z}_1, z_2)$. Now we can rewrite $P(y|\hat{z}_1, z_2, x')$ as $P(y|z_1, z_2, x')$ using Rule 2. $P(x'|\hat{z}_1, z_2)$ can be rewritten as $P(x'|z_2)$ using Rule 3, since $Z_1$ is a child of $X$ and the graph is acyclic. So, the query can be rewritten as $\sum_{z_1, z_2} \sum_{x'} P(y|z_1, z_2, x') P(x'|z_2) P(z_1, z_2|\hat{x})$. $P(z_1, z_2|\hat{x}) = P(z_2|\hat{x}) P(z_1|\hat{x}, z_2)$. Since $Z_2$ consists of non-descendants of $X$, we can rewrite $P(z_2|\hat{x})$ as $P(z_2)$ using Rule 3. Since $Z_2$ blocks all back-door paths from $X$ to $Z_1$, we can rewrite $P(z_1|\hat{x}, z_2)$ as $P(z_1|x, z_2)$ using Rule 2. The entire query can thus be rewritten as $\sum_{z_1, z_2} \sum_{x'} P(y|z_1, z_2, x') P(x'|z_2) P(z_1|x, z_2) P(z_2)$.

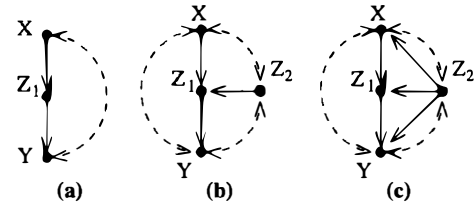

(a)          (b)          (c)

Figure 3: Illustrating Condition 4 of Theorem 2. In **a**, $Z_1$ blocks all directed paths from $X$ to $Y$. The empty set blocks all back-door paths from $Z_1$ to $Y$ in $G_{\overline{X}}$, and all back-door paths from $X$ to $Z_1$ in $G$. In **b** and **c**, $Z_1$ blocks all directed paths from $X$ to $Y$. $Z_2$ blocks all back-door paths from $Z_1$ to $Y$ in $G_{\overline{X}}$, and blocks all back-door paths from $X$ to $Z_1$ in $G$.

It remains to prove that the conditions of Theorem 2 are necessary. This may be shown by contradiction

**Proof Sketch** – We will assume that there exists a query $P(y|\hat{x})$ and a graph $G$ such that : 1. None of the conditions of Theorem 2 hold, and 2. There exists a finite sequence of application of inference rules which removes all hats from the variables in the query. We will show that these two assumptions lead to a contradiction, hence, if all four conditions of Theorem 2 fail, there must not be a finite sequence of inference rules that reduces the query to a hat-free expression.

**Proof Outline** :

**I** $(Y \perp\!\!\!\perp X|Z, W)_{G_{\overline{Z}\,\underline{X}}}$, so rule 2 can be applied to remove the hat from $X$.

    **A** There is a directed path from $Z$ to $Y$
        **1** Cannot add $\hat{z}$ using rule 3
        **2** Cannot add $\hat{z}$ using rule 2
    **B** There is a directed path from $Z$ to $X$
        **1** Cannot remove $\hat{z}$ using rule 2
        **2** Cannot remove $\hat{z}$ using rule 3

**II** $(Y \perp\!\!\!\perp X | Z, W)_{G_{\overline{Z}\,\overline{X(W)}}}$, so rule 3 can be applied to remove $\hat{x}$

    **A** Cannot add $\hat{z}$ using rule 3
    **B** Cannot add $\hat{z}$ using rule 2

Assume that there exists a querry $P(y|\hat{x})$ and a graph $G$ such that none of the conditions of Theorem 2 hold, but the query is still identifiable. Since $P(y|\hat{x})$ is identifiable, there must be some finite sequence of inference rules that removes the hat from $x$. That means, there must be some (possibly empty) set of variables $Z$ and $W$ such that either $(Y \perp\!\!\!\perp X | Z, W)_{G_{\overline{Z}X}}$, so we can reduce $P(y|\hat{x}, \hat{z}, w)$ to $P(y|x, \hat{z}, w)$ via Rule 2, or $(Y \perp\!\!\!\perp X | Z, W)_{G_{\overline{Z}\,\overline{X(W)}}}$, so we can reduce $P(y|\hat{x}, \hat{z}, w)$ to $P(y|\hat{z}, w)$ using Rule 3. We will look at each of these two cases in turn.

**Case I**: First, consider $(Y \perp\!\!\!\perp X | Z, W)_{G_{\overline{Z}X}}$. By assumption, $P(y|\hat{x})$ is identifiable, and the hat is removed from $X$ by an application of Rule 2. This implies a series of rule applications to $P(y|\hat{x})$ which results in $P(y|\hat{x}, \hat{z}, w)$, such that $(Y \perp\!\!\!\perp X | Z, W)_{G_{\overline{Z}X}}$. We will look at the restrictions on $Z, W$ that are imposed by both the failure of the conditions of Theorem 2 to hold, and the assumption that $P(y|\hat{x})$ can be transformed to $P(y|\hat{x}, \hat{z}, w)$ by a series of rule applications. We will also make the assumption that $Z$ and $W$ are minimal. If they are not, then there exists a minimal $Z'$ and $W'$, in which superfluous nodes are removed, which would also work – so proving that no minimal $Z$ and $W$ exist implies that no $Z$ or $W$ exist.

If $(Y \perp\!\!\!\perp X | Z, W)_{G_{\underline{X}}}$, then there would exist a blockable back-door path, and Condition 3 of Theorem 2 would have held. We also know $(Y \perp\!\!\!\perp X | Z, W)_{G_{\overline{Z}X}}$, by assumption. These two independence assertions imply that $Z$ conducts a back-door path that is not blocked by $W$. That is, there is a back-door path between $X$ and $Y$ that has a head-to-head junction in $Z$. Each element of $Z$ must also block a back-door path from $X$ to $Y$, since $Z$ is minimal. This implies that there is a directed path from $Z$ to $X$, or from $Z$ to $Y$ (Figure 4):

Proof that there is a directed path from $Z$ to $X$ or from $Z$ to $Y$ :

Since we know that $Z$ must block a back-door path from $X$ to $Y$, there must be a path from $Z$ to $X$ or from $Z$ to $Y$ that starts in an arrow that is incident away from $Z$. All of the head-to-head junctions along this path must either be in $W$ or have descendants in $W$. If there are no such head-to-head junction paths, then there is a directed path from $Z$ to $X$ or $Y$. If there is a head-to-head junction, then consider the $W$ that unblocks this junction. This $W$ must itself block a back-door path from $X$ to $Y$. So, there must be a path from $W$ to either $X$ or $Y$ that starts with an arc incident away from $W$. This path is either a directed path from $W$ to $X$ or $Y$, or has a head-to-head junction that is also a member of $W$, or is an ancestor of a member of $W$. Since the graph is acyclic, there must eventually be a $W$ that has a directed path to $X$ or $Y$ that is a descendant of $Z$ – thus there is a directed path from $Z$ to either $X$ or $Y$.

We now look at the two cases; Case 1A: a directed path exists from $Z$ to $Y$, Case 1B: a directed path exists from $Z$ to $X$

**Case IA**: A directed path exists from $Z$ to $Y$.

By our assumption, there must be a sequence of rules to transform $P(y|\hat{x})$ to $P(y|\hat{x}, \hat{z}, w)$. There are two ways to add $\hat{z}$ to this expression – either directly using Rule 3, or by first conditioning on $Z$ and then adding a hat to it by using rule 2.

**Case IA1**: First we look at using Rule 3. If there is a directed path from $Z$ to $Y$ (Figure 4a), then $(Y \not\perp\!\!\!\perp Z | X)_{G_{\overline{X}\,\overline{Z}}}$. No element of $W$ can block this path from $Y$ to $Z$, since that would require $W$ to be a descendant of $Z$, and $(Y \not\perp\!\!\!\perp Z | X, W)_{G_{\overline{X}}}$. So Rule 3 cannot be invoked to add $\hat{z}$ to the expression.

**Case IA2**: We need to first condition on $Z$, and then add the hat to it using Rule 2. In order for us to add the hat to $Z$ using rule 2, there needs to be a $W'$ such that $(Y \perp\!\!\!\perp Z | W', X)_{G_{\overline{X}\underline{Z}}}$. Above, we proved that given our assumptions, there must be an unblocked path from $Y$ to $X$ that has a head-to-head junction at $Z$, and no member of $W$ that blocks it, so $W' \not\subseteq W$. If we condition on a $W'$ that allows us to add the hat to $Z$, we must then remove it to obtain $P(y|\hat{x}, \hat{z}, w)$ so that we can remove the hat from $x$. However, we are not able to remove this $W'$. We cannot remove $W'$ using rule 1, since $(Y \not\perp\!\!\!\perp W'|X, Z, W)_{G_{\overline{X}\,\overline{Z}}}$, and if we add some $W''$ that $d$-separates $Y$ from $W'$, then we would not be able to remove $W''$.

Thus we cannot add $\hat{z}$ to $P(y|\hat{x})$ by first conditioning on $Z$ and then adding a hat to it using Rule 2 if there is a directed path from $Z$ to $Y$.

**Case IB**: A directed path exists from $Z$ to $X$.

If there is a directed path from $Z$ to $X$ (Figure 4b), we will assume that we can can add $\hat{z}$ to $P(y|\hat{x})$ to get $P(y|\hat{x}, \hat{z})$, and condition on $W$ to get $P(y|\hat{x}, \hat{z}, w)$. We can then remove the hat from $x$ using Rule 2 to get $P(y|x, \hat{z}, w)$.

Now we will prove that there is no way to remove $\hat{z}$ from the expression. Since there is a back-door path from $X$ to $Y$ that has a head-to-head junction at $Z$, there must be an back-door path from $Z$ to $Y$.

**Case IB1**: If we could remove the hat from $Z$ using Rule 2, then we could block the back door path from $Z$ to $Y$ – and hence we could block the back-door path from $X$ to $Y$, and Condition 3 would have held.

**Case IB2**: If we could remove $\hat{z}$ directly using Rule 3, then there would have to be some set of nodes that blocked the directed path from $Z$ to $X$, and both $(Y \perp\!\!\!\perp X|Z,W)_{G_{\overline{Z}\underline{X}}}$ and $(Y \not\!\perp\!\!\!\perp X|Z,W)_{G_{\underline{X}}}$ would not be true.

Thus we cannot remove all the hats from the expression by removing the hat from $X$ using Rule 2.

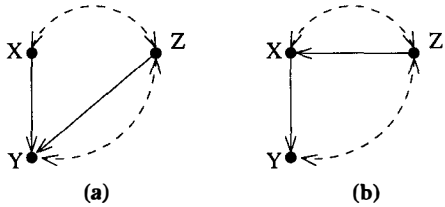

Figure 4: Using Rule 2 to remove the hat from $X$ when the criterion fails. Since $Z$ is necessary, there must be a directed path from : **a** $Z$ to $Y$ or **b** $Z$ to $X$.

**Case II**: Now consider $(Y \perp\!\!\!\perp X|Z,W)_{G_{\overline{Z}\;\overline{X(W)}}}$. We will try to find a set of rule applications that transforms $P(y|\hat{x})$ into $P(y|\hat{x},\hat{z},w)$ when none of the conditions of Theorem 2 hold. $Z$ must block all directed paths from $X$ to $Y$. If it did not, then $W$ would have to block a directed path, but that would make $W$ a descendant of $X$, so $X(W) = \emptyset$, and thus $(Y \perp\!\!\!\perp X|Z,W)_{G_{\overline{Z}}}$, and above we proved that this could not happen if any of the conditions of Theorem 2 held. There are two ways to add $\hat{z}$ to $P(y|\hat{x})$ – either directly using Rule 3, or by conditioning on $Z$ and then adding a hat to it using Rule 2. We will look at each of these in turn.

**Case IIA**: First, we will try to add $\hat{z}$ directly by using Rule 3. To do this, there must be some $W$ such that $(Y \perp\!\!\!\perp Z|W,X)_{G_{\overline{X}\;\overline{Z(W)}}}$. Since there is a directed path from $Z$ to $Y$, $W$ must be a descendant of $Z$, thus $(Y \perp\!\!\!\perp Z|W,X)_{G_{\overline{X}}}$. So, $W$ blocks all back-door paths between $Z$ and $Y$ in $G_{\overline{X}}$. Once $\hat{x}$ has been removed from $P(y|\hat{x},\hat{z},w)$ to obtain $P(y|\hat{z},w)$, we need to remove $\hat{z}$, or remove the hat from $Z$. We cannot remove the hat from $Z$ directly by using rule 3, since $(Y \not\!\perp\!\!\!\perp Z|W,X)_{G_{\overline{Z(W)}}}$, because $Z(W) = \emptyset$, and there is a back door path from $Z$ to $Y$ through $X$. If we could remove the hat from $Z$ by using rule 2, then Condition 4 would have held. So, we cannot add $\hat{z}$ directly using Rule 3 if any of the conditions of Theorem 2 hold.

**Case IIB**: Next, we will try to condition on $Z$, and then add a hat to it using Rule 2. However, if this was possible then there would have to be a $W$ that blocked back door paths between $X$ and $Z$, and blocked back-door paths between $Z$ and $Y$ – and then Condition 4 would have held.

Thus, if none of the conditions of Theorem 2 hold, the query must not be identifiable.

**Remark**: The criterion in Theorem 2 is complete only if the inference rules themselves are complete. The appendix shows that the graphical conditions which license each of the rules in Theorem 1 are the tightest possible. However, the possibility still remains that some strange exchange of hatted and non-hatted variables would be licensed by some graph, and not reachable by successive applications of Rules 1-3.

## 4 Remarks on Efficiency

In implementing Theorem 2 as a systematic method of determining identifiability, Conditions 3 and 4 would seem to require an exhaustive search. To prove that Condition 3 does not hold, for instance, we need to prove that no blocking set $B$ can exist. Fortunately, the following theorems allow us to significantly prune the search space, so as to render the test tractable.

**Theorem 3** *If, for one minimal set $B_i$, $P(b_i|\hat{x})$ is identifiable, then for any other minimal set $B_j$, $P(b_j|\hat{x})$ is also identifiable.*

Theorem 3 allows us to test Condition 3 with a single minimal blocking set $B$. If $B$ meets the requirements for Condition 3, then the query is identifiable, otherwise Condition 3 cannot be satisfied.

**Theorem 4** *Let $Y_1$ and $Y_2$ be two subsets of nodes such that either no nodes $Y_1$ are descendants of $X$, or all nodes $Y_1$ and $Y_2$ are descendants of $X$ and all nodes $Y_1$ are nondescendants of $Y_2$. Then, there exists a reducing sequence for $P(y_1, y_2|\hat{x})$ (per Corollary 1) if and only if there are reducing sequences for both $P(y_1|\hat{x})$ and $P(y_2|\hat{x}, y_1)$.*

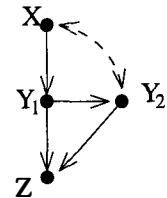

Figure 5: Theorem 2 Ensures a reducing sequence for $P(y_2|\hat{x}, y_1)$ and $P(y_1|\hat{x})$, although none exists for $P(y_1|\hat{x}, y_2)$.

It is possible that $P(y_1, y_2|\hat{x})$ will pass the test in Theorem 2 if we apply the procedure to both $P(y_2|\hat{x}, y_1)$ and $P(y_2|\hat{x})$, but if we try to apply the test to $P(y_1|\hat{x}, y_2)$, we will not find a reducing sequence of rules. Figure 5 shows just such an example. Theorem 4, however, guarantees that, if there is a reducing sequence for $P(y_1, y_2|\hat{x})$ then we should always be able to

find such a sequence for both $P(y_1|\hat{x})$ and $P(y_2|\hat{x}, y_1)$ by proper choice of $Y_1$.

**Theorem 5** *If there exists a set $Z_1$ that meets all of the requirements for $Z_1$ in Condition 4, then the set consisting of the children of $X$ intersected with the ancestors of $Y$ will also meet all of the requirements for $Z_1$ in Condition 4.*

Theorem 5 removes the need to search for $Z_1$ in Condition 4 of Theorem 2.

**Proof of Theorem 3**

If, for one minimal set $B_i$, $P(b_i|\hat{x})$ is identifiable, then for any other minimal set $B_j$, $P(b_j|\hat{x})$ is also identifiable.

We will use the following lemma :

**Lemma 1** *If the query $P(y|\hat{x})$ is identifiable, and a set of nodes $Z$ lies on a directed path from $X$ to $Y$, then the query $P(z|\hat{x})$ is identifiable.*

**Proof** (By Contradiction) :

Assume that there is a minimal set $B$ such that $(Y \perp\!\!\!\perp X|B)_{G_{\underline{X}}}$ and the query $P(b|\hat{x})$ is identifiable. Assume that there is another minimal set $K$ such that $(K \perp\!\!\!\perp X|B)_{G_{\underline{X}}}$, and the query $P(k|\hat{x})$ is not identifiable.

Consider all (undirected) paths from $X$ to $Y$ in $G_{\underline{X}}$. Every element of $B$ and $K$ must lie along one of these paths, since the sets are minimal. In addition, at least one member of $K$ must be a descendant of $X$, otherwise $P(k|\hat{x})$ would be identifiable. In fact, any member of $K$ that is a descendant of $X$ needs to lie on a directed path from $X$ to $Y$.

Proof that any member of $K$ that is a descendant of $X$ lies on a directed path from $X$ to $Y$ :

If a member $K_1$ of $K$ was a descendant of $X$ but did not lie on a directed path from $X$ to $Y$, then there must be a head-to-head junction along the path from $K_1$ to $Y$. This path would have to be unblocked by some other member $K_2$ of $K$. Since $K$ is minimal, there must be some unblocked path from some descendant of $K_2$ to $Y$ that $K$ blocks. This implies that there is either a directed path from one of the descendants of $K_2$ to $Y$, which would make $K_1$ an ancestor of $Y$, or there must be a head-to-head junction on the path from $K_2$ to $Y$ that is unblocked by some other member $K_3$ of $K$. There is either an infinite series of $K$s between $K_1$ and $Y$, or else a directed path from $K_1$ to $Y$ (see Figure 6).

Let $K'$ be the subset of $K$ that lies on a directed path from $X$ to $Y$, and let $K'' = K \setminus K'$. We know that $P(k|\hat{x}) = P(k'|\hat{x}, k'') * P(k''|\hat{x})$, and that $P(k''|\hat{x}) = P(k'')$. So, $P(k'|\hat{x}, k'')$ must not be identifiable. Since $K$ is minimal, $K'$ must block some back-door path,

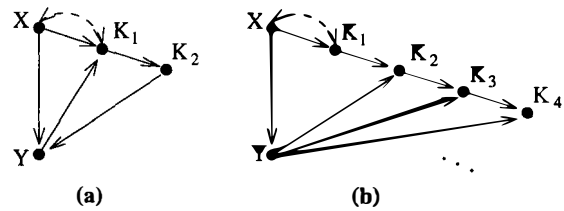

Figure 6: If a member of $K$ blocks a back-door path from $X$ to $Y$, and is a descendant of $X$, then it is also an ancestor of $Y$.

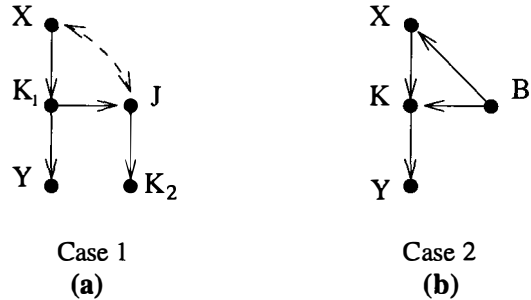

Case 1                Case 2
(a)                   (b)

Figure 7: Examples of the two cases for $K'$

and that back-door path must also be blocked by some member of $B$. There are two possibilities : either the path that $K'$ blocks has a head-to-head junction that is not unblocked by $B$, or there is some member of $B$ which blocks the same path. An illustration of these two cases can be found in Figure 7. Looking at each of these in turn :

**Case 1** : There is a head-to-head junction that is not unblocked in $B$, but is unblocked in $K$. Call this junction $J$. Since $K$ is minimal, the element of $K$ that unblocks this path (either equal to $J$, or one of $J$'s descendants) must lie on some unblocked path from $Y$ to $X$ in $G_{\underline{X}}$. If this is the case, then there must be an unblocked path through $J$'s descendants that also goes through $J$, which means there must be some element $B'$ of $B$ that blocks the path between $J$ and $X$ in $G_{\underline{X}}$. (See Figure 8) We can condition and sum over this $B'$ to get :

$$P(k'|\hat{x}, k'') = \sum_{b'} P(k'|\hat{x}, k'', b') * P(b'|\hat{x}, k'')$$
$$= \sum_{b'} P(k'|x, k'', b') * P(b'|\hat{x}, k'')$$

by using Rule 2. So the query $P(b'|\hat{x}, k'')$ must not be identifiable. Thus $B'$ must be a descendant of $X$, otherwise $P(b'|\hat{x}, k'') = P(b'|k'')$. So, $P(b'|\hat{x})$ is identifiable, but $P(b'|\hat{x}, k'')$ is not. Therefore, $K''$ must disallow the blocking of a back-door path from $X$ to $B'$. Then there must be a back-door path from $X$ to $B'$ that has a head-to-head junction, and that junction has a descendant in $K''$, but not in $B$. This is

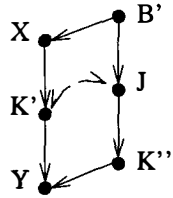

Figure 8: There must exist a member $B'$ of $B$ which blocks the back-door path from $X$ to $J$.

impossible – since $K$ is minimal, the descendant of the head-to-head junction must block a back-door path from $X$ to $Y$. $B$ must block that same path, meaning the path from $X$ to $B'$ was unblocked by $B$ as well as by $K''$.

**Case 2**: There is a member $B'$ of $B$ that blocks the same back-door path as $K'$. The path could be blocked by $B'$ either between $X$ and $K'$, or between $K'$ and $Y$ (See Figure 9). If the path is blocked by $B'$ between $X$ and $K'$, we have the same contradiction as in Case 1 above. If it is blocked by $B'$ between $K'$ and $Y$, then $B'$ lies on a directed path from $X$ to $Y$. From Lemma 1, we know that $P(k'|\hat{x})$ must be identifiable. That means that $K''$ must disallow either Condition 3 or 4 of the Theorem 2. If it blocks Condition 3, then $K''$ must conduct a back-door path from $X$ to $K'$. That means that there must be some member of $K''$ that is at a head-to-head junction along a path from $K'$ to $X$ in $G_{\underline{X}}$, or is a descendant of that junction. Using the same argument as above, since $K$ is minimal, the path blocked by $K''$ must also be blocked by $B$, and thus the head-to-head junction must be unblocked by $B$ as well. Any unblockable back-door path from $X$ to $K'$ will also be an unblockable back-door path from $X$ to $B'$, since $B'$ is a direct descendant of $K'$. However, we know that there cannot be a back-door path from $X$ to $B'$ that is unblockable when we condition on $B$, – thus there cannot be a back-door path from $X$ to $K'$ that is unblockable when we condition on $K''$

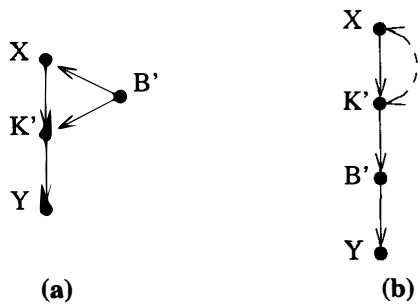

Figure 9: $B$ can be either : **a** between $X$ and $K'$, or **b** between $K'$ and $Y$.

If $K''$ disallows Condition 4, then there must be some other set of nodes $R$ that blocks every directed path from $X$ to $K'$. $K''$ must unblock a back-door path from $X$ to $R$ or from $R$ to $K'$. As above, if a back-door path from $X$ to $R$ (and thus from $X$ to $K'$) is unblocked by $K''$, it will also be unblocked by $B$. So, $K''$ must unblock a back-door path from $R$ to $K'$. Since $K$ is minimal, there must be a path from a descendant of $R$ to $X$ in $G_{\underline{X}}$. So, there must also be a path from $Y$ to $X$ in $G_{\underline{X}}$ that passes through $R$ and $K'$. Since the back-door path from $X$ to $K'$ must not be blockable (since a blockable back-door path was invalidated above), $B$ must block the path from $Y$ to $K'$. But then there would not be a back-door path from $R$ to $K'$ that is blockable when conditioning on $B$, but unblockable when conditioning on $K''$.

So, if any minimal set $B$ blocks all back-door paths from $X$ to $Y$, and the query $P(b|\hat{x})$ is identifiable, then if any other minimal set $K$ also blocks all back-door paths from $X$ to $Y$, then $P(k|\hat{x})$ must also be identifiable.

**Proof of Lemma 1** :

If the query $P(y|\hat{x})$ is identifiable, and a set of nodes $Z$ lies on a directed path from $X$ to $Y$, then the query $P(z|\hat{x})$ must be identifiable.

If the query $P(y|\hat{x})$ is identifiable, one of the four conditions of Theorem 2 must have been satisfied. Looking at each in turn :

Condition 2 - If there is no path from $Y$ to $X$ in $G_{\underline{X}}$, then there cannot be a path from any of $Y$'s ancestors to $X$ in $G_{\underline{X}}$, since any path from $X$ to $Z$ would be part of a path from $X$ to $Y$.

Condition 1 – If there is no directed path from $Y$ to $X$, then there cannot be a $Z$ that lies along a directed path from $Y$ to $X$, and the lemma is trivially true.

Condition 3 – If there is a set $B$ that blocks all back-door paths from $X$ to $Y$, then : any back-door path from $X$ to $Z$ will also be a back-door path from $X$ to $Y$. $B$ must block this back-door path from $X$ and $Y$. If $B$ blocks the path between $X$ and $Z$, then $B$ also blocks the back-door path from $X$ to $Z$, and the query $P(z|\hat{x})$ is identifiable. If $B$ blocks the path between $Z$ and $Y$, then we can use the fact that the query $P(b|\hat{x})$ must be identifiable. If $P(b|\hat{x})$ is identifiable by Condition 4, then $P(z|\hat{x})$ must also be identifiable by Condition 4, since the variables that meet the specifications for $Z_1$ in condition 4 for $P(b|\hat{x})$ will also meet the specifications for $Z_1$ in Condition 4 for $P(z|\hat{x})$. If $P(b|\hat{x})$ is identifiable by Condition 3, then there is some $B'$ that blocks the back-door path from $X$ to $B$, either between $X$ and $Z$, in which case $P(z|\hat{x})$ is identifiable, or between $Z$ and $B'$. Since there are a finite number of links between $Z$ and $Y$, eventually the back-door path from $X$ to $Z$ must be blocked, and the query $P(z|\hat{x})$ is identifiable.

Condition 4 – If there exists a set $Z_1$ and $Z_2$, $Z$ can either come before $Z_1$ or after $Z_1$. If it comes after $Z_1$, then the conditions that held for $Y$ will also hold for $Z$, and the query will be identifiable. If it comes before $Z_1$, then $\{Z_1, Z_2\}$ will block all back-door paths from $X$ to $Z$, and the query will also be identifiable.

**Proof of Theorem 4**

(By Contradiction). Let $Y_1$ and $Y_2$ be two subsets of nodes such that either no nodes $Y_1$ are descendants of $X$, or all nodes $Y_1$ and $Y_2$ are decendants of $X$ and all nodes $Y_1$ are nondescendants of $Y_2$. Assume that there exists a reducing sequence for both $P(y_2|\hat{x})$ and $P(y_1|\hat{x}, y_2)$, but not for $P(y_2|\hat{x}, y_1)$. There are 3 possible cases :

Case 1: $Y_1$ and $Y_2$ are both non-descendants of $X$. In this case, $P(y_1|\hat{x}, y_2) = P(y_1|y_2)$ and is thus identifiable.

Case 2: $Y_2$ is a descendant of $X$, but $Y_1$ is not. In this case, $Y_1$ must unblock a back-door path from $X$ to $Y_1$ which cannot be blocked by conditioning on other variables. But if this is the case, then there must be an unblockable back-door path from $X$ to $Y_1$. Since $Y_1$ is a descendant of $X$, that would make $P(y_1|\hat{x}, y_2)$ unidentifiable.

Case 3: $Y_1$ and $Y_2$ are both descendants of $X$. $Y_1$ cannot unblock a back-door path from $X$ to $Y_2$ since $Y_1$ is an ancestor of $Y_2$. Thus $P(y_2|\hat{x})$ must be unidentifiable, and thus $P(y_2|\hat{x}, y_1)$ is also unidentifiable.

**Proof of Theorem 5**

If there exists a set $Z_1$ that meets all of the requirements for $Z_1$ in Condition 4 of Theorem 2, then the set consisting of the direct descendants of $X$ intersected with the ancestors of $Y$ will also meet all of the requirements for $Z_1$ in Condition 4.

Assume that there exists some set $Z_1$, which does not consist entirely of children of $X$, such that $Z_1$ blocks all directed paths from $X$ to $Y$, and there also exists a set $Z_2$ that blocks all back-door paths from $X$ to $Z_1$ in $G$, and all back-door paths from $Z_1$ to $Y$ in $G_{\overline{X}}$. Let $Z_1'$ be the intersection of the children of $X$ with the ancestors of $Y$. $Z_1'$ clearly blocks all directed paths from $X$ to $Y$. Any back-door path from $X$ to $Z_1'$ must also be part of a back-door path from $X$ to some member of $Z_1$, since every member of $Z_1$ must be either a member of $Z_1'$ or a descendant of some member of $Z_1'$. Since $Z_2$ consists of non-descendants of $X$, $Z_2$ must block all back-door paths from $X$ to $Z_1$ between $X$ and $Z_1'$ – so $Z_2$ also blocks all back-door paths from $X$ to $Z_1'$. Similarly, all back-door paths from $Z_1'$ to $Y$ are also part of back-door paths from $Z_1$ to $Y$, which are also blocked by $Z_2$.

## 5 Complexity Analysis

Using the results of Section 4, we can show that the identifiability test provided by Theorem 2 can be implemented in polynomial time. We will show that each of the four conditions in Theorem 2 can be tested in polynomial time.

1. Since $d$-separation can be determined in time $O(V + E)$, Condition 2 can be tested in polynomial time

2. Again, since $d$-separation can be determined in time $O(V + E)$, Condition 1 can be tested in polynomial time

3. Theorem 3 allows us to test a single minimal blocking set to determine if Condition 3 holds. Thus, we need to find a minimal blocking set between two variables. This can be done in polynomial time as follows :

    (a) Set $R_1 = X$ and $R_2 = \mathbf{pa}_X$
    (b) For each $r \in R_2$ that has a confounding (two-headed) link to a member of $R_1$, remove $r$ from $R_2$, add $r$'s parents to $R_2$, and add $r$ to $R_1$
    (c) If $R_2 \cap Y \neq \emptyset$, return FAIL.
    (d) Set $R_3 = Y$ and $R_4 = \mathbf{pa}_Y$
    (e) For each $r \in R_4$ that has a confounding (two-headed) link to a member of $R_3$, remove $r$ from $R_4$, add $r$'s parents to $R_4$, and add $r$ to $R_3$
    (f) If $R_4 \cap X \neq \emptyset$, return FAIL.
    (g) Set $B = R_2 \cup R_4$
    (h) If $(Y \not\!\perp\!\!\!\perp X | B)$, return Fail.
    (i) For each member $b$ of $B$, if $(Y \perp\!\!\!\perp X | B \setminus b)$, remove $b$ from $B$.
    (j) If anything was removed from $B$ in step i, Go To step i.
    (k) Return $B$.

4. To test condition 4, we need to find a set of variables $Z_1$ and $Z_2$. Theorem 5 gives us a constant-time method to choose $Z_1$. To find $Z_2$, we need only find a blocking set that is not a descendant of $X$. We can do this by labeling the descendants of $X$ "unobservable" and using the method above for finding a minimal blocking set.

## 6 Conclusions

Although this paper focuses on the task of testing identifiability, it should be emphasized that whenever identifiability is confirmed by the test, a closed-form formula for $P(y|\hat{x})$ can easily be assembled using the inference rules invoked in the test. Details of this construction are not shown explicitly in the paper, but can be found in [Pearl, 1994a, Pearl, 1994b]. In view of this construction, we now have a polynomial-time method of assessing the ramification of actions, given a qualitative causal diagram together with a probability distribution on a set of observed variables. As mentioned in the introduction, the main application of this result in AI settings lies in enabling one agent to learn to act by passively observing the performance of other acting agents, even in cases where the actions of those other agents are predicated on factors that are not visible to the learner. If the learner is permitted to act as well as observe, the task becomes much easier of course, because the topology of the diagram can also be inferred, at least partially. Immediate applications to cause effect analysis of nonexperimental data in the social and medical sciences are discussed in [Pearl, 1994a].

## 7 Appendix

The conditions which license each of the rules of Theorem 1 are the tightest possible :

$$(Y \perp\!\!\!\perp Z|X,W)_{G_{\overline{X}}} \quad \text{if} \quad P(y|\hat{x}, z, w) = P(y|\hat{x}, z)$$

Since the $d$-separation condition is valid for any recursive model, including the submodel represented by $G_{\overline{X}}$, the conditional independence $P(y|\hat{x}, z, w) = P(y|\hat{x}, z)$ implies $(Y \perp\!\!\!\perp Z|X,W)_{G_{\overline{X}}}$.

$$(Y \perp\!\!\!\perp Z|X,W)_{G_{\overline{X}\underline{Z}}} \quad \text{if} \quad P(y|\hat{x}, \hat{z}, w) = P(y|\hat{x}, z, w)$$

Consider the augmented diagram $G'$ that has the intervention arcs $F_Z \to Z$ added. $P(y|\hat{x}, \hat{z}, w) = P(y|\hat{x}, z, w)$ implies that $(Y \perp\!\!\!\perp F_Z|X,Z,W)_{G'_{\overline{X}}}$. If there is a path from $Y$ to $Z$ that is unblocked by $\{X,W\}$ in $G_{\overline{X}}$, it must not end in an arrow incident to $Z$, otherwise $(Y \perp\!\!\!\perp F_Z|X,Z,W)_{G'_{\overline{X}}}$ would not hold. Since every path from $Y$ to $Z$ that is not blocked by $\{X,W\}$ in $G_{\overline{X}}$ must pass through an arrow leaving $Z$, $(Y \perp\!\!\!\perp Z|X,W)_{G_{\overline{X}\underline{Z}}}$.

$$(Y \perp\!\!\!\perp Z|X,W)_{G_{\overline{X}\,\overline{Z(W)}}} \quad \text{if} \quad P(y|\hat{x}, \hat{z}, w) = P(y|\hat{x}, w)$$

Again consider $G'$ with intervention arcs $F_Z \to Z$ added. $P(y|\hat{x}, \hat{z}, w) = P(y|\hat{x}, w)$ implies that $(Y \perp\!\!\!\perp F_Z|X,W)_{G'_{\overline{X}}}$. So, any path from $Z$ to $Y$ that is not blocked by $\{X,W\}$ in $G_{\overline{X}}$ must end in an arrow pointing to $Z$, otherwise $(Y \perp\!\!\!\perp F_Z|X,W)_{G'_{\overline{X}}}$ would not hold. In addition, if there is a path from some $Z'$ of $Z$ to $Y$ that does end in an arrow pointing to $Z'$, then $W$ must not be a descendant of $Z'$, otherwise $(Y \perp\!\!\!\perp F_Z|X,W)_{G'_{\overline{X}}}$ would not hold. Thus the only paths from $Y$ to $Z$ must end in an arrow pointing at $Z$, and must end in some member of $Z(W)$. Thus, $(Y \perp\!\!\!\perp Z|X,W)_{G_{\overline{X}\,\overline{Z(W)}}}$.

### Acknowledgements

The research was partially supported by Air Force grant #AFOSR/F496209410173, NSF grant #IRI-9420306, and Rockwell/Northrop Micro grant #94-100. Max Chickering provided much valuable assistance with the proofs in this document. We thank James Robins for commenting on the first draft of this paper.